\documentclass[journal]{IEEEtran}
\usepackage{amsmath,amsfonts}
\usepackage{algorithmic}
\usepackage{algorithm}
\usepackage{array}
\usepackage[caption=false,font=normalsize,labelfont=sf,textfont=sf]{subfig}
\usepackage{textcomp}
\usepackage{stfloats}
\usepackage{url}
\usepackage{verbatim}
\usepackage{graphicx}
\usepackage[capitalise]{cleveref}
\usepackage[numbers]{natbib}
\usepackage[printonlyused]{acronym}
\usepackage{siunitx}
\DeclareMathAlphabet{\mathpzc}{OT1}{pzc}{m}{it}
\usepackage{ctable}
\usepackage{multirow}

\newacro{MDP}{markov decision process}
\newacro{RL}{reinforcement learning}
\newacro{MARL}{multi-agent reinforcement learning}
\newacro{MAS}{multi-agent systems}
\newacro{GEE}{global entity encoder}
\newacro{MRMG}{multi-robot multi-goal}
\newacro{CTDE}{centralized training decentralized execution}
\newacro{DoF}{degree of freedoms}
\newacro{GNN}{graph neural networks}
\newacro{Dec-POMDP}{decentralized partially observable Markov decision process}
\newacro{MPC}{model predictive control}
\newacro{TAMP}{task assignment and motion planning}
\newacro{MAF}{multi-agent foraging}

\begin{document}
\title{
Solving Multi-Entity Robotic Problems Using Permutation Invariant Neural Networks

}

\author{Tianxu An*$^{1}$, Joonho Lee*$^{1}$, Marko Bjelonic$^{2}$, Flavio De Vincenti$^{3}$, Marco Hutter$^{1}$

\thanks{* \textit{Tianxu An and Joonho Lee contributed equally to this work.} (\textit{Corresponding author: Joonho Lee}).}

\thanks{$^{1}$ The authors are with Robotic Systems Lab; ETH Zurich, Switzerland
}
\thanks{$^{2}$ The author was with Robotic Systems Lab; ETH Zurich, Switzerland. The author is now with Swiss-Mile Robotics AG, Switzerland.
}
\thanks{$^{3}$ The author is with Computational Robotics Lab; ETH Zurich, Switzerland
}
}



\maketitle

\begin{abstract}
Challenges in real-world robotic applications often stem from managing multiple, dynamically varying entities such as neighboring robots, manipulable objects, and navigation goals. Existing multi-agent control strategies face scalability limitations, struggling to handle arbitrary numbers of entities. Additionally, they often rely on engineered heuristics for assigning entities among agents. We propose a data driven approach to address these limitations by introducing a decentralized control system using neural network policies trained in simulation. Leveraging permutation invariant neural network architectures and model-free reinforcement learning, our approach allows control agents to autonomously determine the relative importance of different entities without being biased by ordering or limited by a fixed capacity. We validate our approach through both simulations and real-world experiments involving multiple wheeled-legged quadrupedal robots, demonstrating their collaborative control capabilities. We prove the effectiveness of our architectural choice through experiments with three exemplary multi-entity problems. Our analysis underscores the pivotal role of the end-to-end trained permutation invariant encoders in achieving scalability and improving the task performance in multi-object manipulation or multi-goal navigation problems. The adaptability of our policy is further evidenced by its ability to manage varying numbers of entities in a zero-shot manner, showcasing near-optimal autonomous task distribution and collision avoidance behaviors.
\end{abstract}


\section{Introduction}


\IEEEPARstart{H}{umans} routinely deal with multi-entity problems in daily life. For example, human workers collaborate with co-workers to construct structures, or a group of friends splits up to find various products in a grocery store.
In such scenarios, individuals naturally decide where to work within the site, with whom to collaborate, and which object or sub-task to prioritize.
Can robots have similar capabilities? Despite recent advances that have made robots proficient in tasks like locomotion~\citep{Lee_2020, walk_in_min, Miki_2022}, navigation~\citep{anymal_navigation, smug_nav, sound_nav}, and object manipulation~\citep{dense_bin_packing, loose_packing, practical_multi_obj}, existing approaches cannot be readily applied to collaborative multi-entity tasks.
The majority of literature focuses on solving single-robot problems, leaving a gap in addressing the complexities of multi-agent collaboration and coordination in real-world scenarios.

\begin{figure}[!t]
    \centering
    \includegraphics[width=\columnwidth]{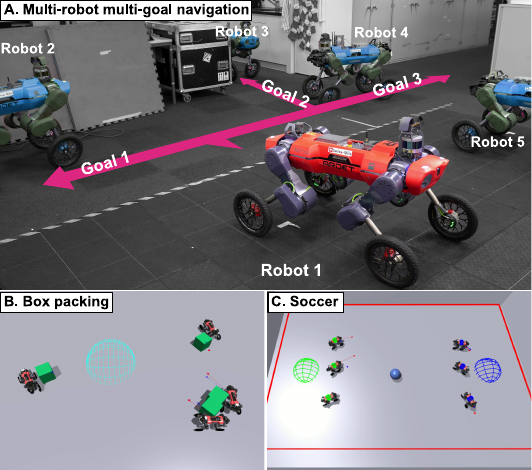}
    \caption{Multi-entity problems studied in this work. (A) Robot 1 is given multiple goals to visit while interacting with other robots.
(B) Multiple robots packing multiple boxes into the goal region. (C) Soccer. }
    \label{fig:opener}
\end{figure}

Our goal is to develop a framework to tackle multi-entity problems involving multiple mobile robots, as illustrated in \cref{fig:opener}.
To explore this problem, we define three multi-entity problems that involve four entity types: collaborators, opponents, navigation goals, and objects to manipulate.
Specifically, \cref{fig:opener}A shows a \ac{MRMG} navigation problem where robots are given multiple navigation goals, \cref{fig:opener}B illustrates box packing problem where robots have to move boxes to the packing site in the center, and \cref{fig:opener}C shows a soccer game.
Each task episode may have a different number of entities of each type. We aim to develop a control strategy that can effectively handle flexible number of entities and generalize to new scenarios.



Existing approaches to controlling \ac{MAS} struggle at tackling two main challenges.
First, scaling up to flexible numbers of entities. 
Traditional centralized methods~\citep{central_cable-towed_load, commNet, skill_behav_diver, flavio_mpc} struggle at scaling their controllers to higher numbers of entities, because the number of decision variables can grow arbitrarily depending on the number of entities, leading to high computational cost.
Although decentralized controllers can alleviate the computation cost, many of them still focus on tasks with a fixed number of entities. For example, multi-agent controllers in~\citep{vision60_locomotion, legged_payload, sim2real_legged_robots} only accept a fixed number of robots. 
Similarly, multi-agent policy by \citet{hide_seek} limits the maximum numbers of agents and manipulable objects.
Second, prioritization and assignment of goals and objects among agents is not straightforward.
A common practice is to assign goals and objects among agents based on predefined rules or heuristics, such as manually assigning each agent with a specific goal~\citep{multi-arm, cm3, mamg_pick_up_delivery} or assigning objects based on proximity or manual commands~\citep{2_arm_brick_piler, multi_arm_multi_box, mr_robots}.

To tackle these challenges, we develop decentralized \ac{MARL} policies that enable a robot to work with flexible numbers of entities. 
We adopt a decentralized approach to maintain constant inference time, leveraging distributed computational costs regardless of the number of agents. 
Additionally, we opt for a model-free \ac{RL} approach to eliminate hand-engineered heuristics typically present in optimization-based approaches.

We further enhance the performance of our agents when prioritizing and distributing entities in long-horizon tasks by using a permutation-invariant neural network called \ac{GEE}. 
Leveraging an architecture similar to PointNet~\cite{point_net}, \ac{GEE} processes an arbitrary number of state vectors from all entities for each entity type.
This architectural choice ensures that agents can act consistently regardless of the order in which they perceive the entities. In other words, the action remains invariant to the permutations in the input space (i.e., order of entities). 
By eliminating order dependency in the decision making, we enable agents to assess the relative importance of different entities without being biased by ordering. In doing so, we remove the need for engineered heuristics.


We validate our approach on the the previously mentioned multi-entity problems in \cref{fig:opener}. For each problem, we use the same policy to control each robot. The robots with identical policy collaborate with an arbitrary number of other robots across various task configurations.
Our permutation invariant network policy effectively addresses multi-entity tasks and outperforms typical fixed-input encoders.
Simulation experiments show that our approach enables robots to autonomously prioritize entities depending on the situation and engage in collaborative problem solving.

Furthermore, we conduct real-world \ac{MRMG} navigation experiments with two wheeled-legged quadrupedal robots~\citep{bjelonic2019keep}, with communication facilitated by WI-FI and gRPC framework~\citep{grpc}.
Our real-world experiments demonstrate intelligent multi-agent navigation behavior, where robots autonomously distribute tasks and avoid collisions.

Our key contributions are as follows:
\begin{enumerate}
    \item We introduce a scalable control policy for multi-entity robotic problems. Our policy is based on  permutation invariant neural networks, which allows it to handle flexible numbers of different entities. 
    \item We verify the effectiveness of our system through real-world multi-entity navigation experiments with two wheeled-legged quadrupedal robots~\citep{bjelonic2019keep}. We provide important technical details regarding training and hardware implementation for communication.
\end{enumerate}


\section{Related Work}
\subsection{Permutation Invariant Neural Networks}
Permutation invariant neural networks ensure output invariance to permutations of the input data. 
Set inputs are common in the robotics domain and take the form of point clouds, sensory inputs, and neighboring agents' states in multi-agent problems. Various network structures achieve the permutation invariance property. For instance, PointNet~\citep{point_net} and DGCNN~\citep{dgcnn} use pooling operations on point cloud inputs. 
Self-attention encoders~\citep{attention} can also preserve permutation invariant inputs by removing any positional encodings from the input elements~\citep{set_transformer, sensory_transformer}. Another line of work uses Mean-Field Approximation to model large-scale multi-agent states or collective actions, thus maintaining the permutation invariance properties of neighboring agents~\citep{mean_field_RL, MF_PPO}. Likewise, \citet{PIPO} ensure the permutation invariance using explicit constraints. They introduce a decentralized multi-agent navigation algorithm that constrains the network outputs to match those with shuffled orders of neighboring agents. In this work, we adopt the PointNet~\citep{point_net} architecture to ensure permutation invariance of all entities in the same category, such as robots, objects, and goals.

\subsection{Multi-object Manipulation}
Many research tackle the problem of controlling robots to manipulate multiple objects. A large body of work in multi-object manipulation focuses on a single manipulator sequentially packing various objects~\citep{dense_bin_packing, loose_packing, tossing_bot}. 
These methods prioritize identifying individual objects and their visual relationship with the background but de-emphasize the inter-object relationships and the manipulation sequence. \citet{practical_multi_obj} and \citet{multi_obj_gnn} use \ac{GNN} to reason about relationships among variable numbers of objects, enabling a single manipulator arm to stack blocks into multiple configurations by planning a manipulation sequence. 

It became more popular recently to control multiple agents to increase efficiency in multi-object manipulation. \citet{multi_arm_multi_box} propose a differentiable \ac{TAMP} strategy to control multiple manipulator arms to smoothly handover blocks. The authors also experiment on the scenarios with multiple objects. However, the approach uses manual heuristics to decide the manipulation sequence. Another line of work in multi-agent multi-object manipulation is \ac{MAF}~\citep{forMIC, spatial_intention_map}. \ac{MAF} problems closely resemble our box packing task, both of which ask a team of agents to pack arbitrary numbers of loads into one or multiple packing sites. Agents in \ac{MAF} often need to closely interact with collaborators and automatically determine the manipulation sequence of objects.
Similar to \ac{MAF}, in our box packing task, we ask multiple robots to implicitly plan a manipulation sequence that considers all objects and collaborators. This requires each robot to discern the relationship of objects with respect to the target, collaborators, and themselves.

\subsection{Multi-goal Navigation}
When it comes to multi-goal navigation, a robot usually receives all the goal locations at the beginning and then navigates to each one. \citet{multi-obj-search} use a greedy strategy that always navigates the robot to the closest goal in an unknown map, while \citet{smug_nav} propose computing an offline path from a known map without a specific goal-visiting order and following the planned path during execution. Both methods are likely to get stuck if a goal point is unreachable, as common in real-world scenarios, e.g., due to improper goal setup or unexpected map changes.

For \ac{MRMG} navigation, early works separate goal assignment and cooperative path planning into two problems~\citep{cm3, mamg_pick_up_delivery}. More recent works apply \ac{MARL} algorithms to let robots autonomously decide goal assignments and cooperative path planning with neighbor awareness~\citep{spatial_intention_map, ATOC}. 
Similar to this approach, we couple goal assignment and cooperative path planning into a single problem. Our policies are optimized synchronously to prioritize and distribute goals while planning their motion to avoid collisions with neighbors.

\subsection{Competitive RL}
Competitive \ac{RL} adopts self-play to simultaneously improve policies of the training teams and their opponents in a wide range of games, such as hide and seek, Go, soccer, and Sumo~\citep{hide_seek, alphago, soccer_sim, soccer_real, multi_agent_competition}. In our soccer task, we also apply self-play. Unlike soccer policies in~\citep{soccer_sim, soccer_real}, our policies do not restrict the number of players on either team, thanks to the permutation invariant encoder which enable ego robot to accept arbitrary numbers of teammates and opponents.

\subsection{Neighbor Awareness in Multi-Agent Systems}
Most \ac{MAS}s require robots to be aware of their neighbors to facilitate effective collaborative behaviors. Early works consider fully decentralized controllers and attempt to induce neighbor awareness through environment changes~\cite{sim2real_legged_robots, iql, IA2C, IPPO}. However, these approaches are limited when agents work closely, as the agents can collide with each other due to the lack of mutual observability. In contrast, centralized strategies guarantee global neighbor awareness, but they suffer from high computational burdens and scalability issues~\citep{central_cable-towed_load, commNet, skill_behav_diver, flavio_mpc}. To mitigate these problems while incentivizing collaboration, some researchers adopt the \ac{CTDE} paradigm~\citep{MAPPO, MADDPG, VDN, QMIX, QPLEX, MACPF, CADP}. In this approach, the agents do not actively communicate at execution time, and all emerging collaborations are the result of a centralized training in physics-based simulators.

In recent times, there has been an increase in the use of wireless networks to enable communication between agents for enhanced collaboration during the execution. Optimization-based controllers have utilized these communication channels to control a swarm of drones for navigation in the wild~\citep{drone_swarm_wild}. However, optimization-based methods are often computationally expensive and may require heuristics, resulting in sub-optimal performance. Therefore, many researchers have combined communication networks with decentralized \ac{RL} policies to achieve faster execution times and more robust behaviors.

Early approaches using \ac{RL} endow agents with an auxiliary neural network for generating communication messages and deciding whether to share them with others~\citep{DIAL, IC3Net}. However, in these approaches, receiving agents passively process the messages either discretely or by aggregating their mean from all broadcasting neighbors. This approach can be problematic because each agent essentially treats all neighboring agents as equally important, often resulting in inundation with irrelevant information while potentially overlooking crucial neighbors. ATOC~\citep{ATOC} tackles this issue by employing an Attention Network to determine if an agent should initiate communication, while a bi-directional LSTM network handles shared messages from collaborators. \citet{multi-arm} employ a similar approach using LSTM for message aggregation to enable a variable number of manipulators to converge on a joint end-effector state using a decentralized control policy for each. Nevertheless, using LSTM as the aggregator poses challenges as neighboring agents are not permutation invariant, necessitating the reordering of neighbors based on hand-engineered heuristics. This can lead to earlier neighbors' states being forgotten, potentially limiting the ego agent's ability to learn from the most important neighbors.

\citet{MAGIC} introduces a Graph-Attention based approach aimed at enhancing multi-agent coordination. Initially, a central planner learns a graph representing communication directions at each policy step. Subsequently, an Attention network is employed for each agent to prioritize messages from neighbors, considering incoming communication directions, including messages from the agent itself. While this mechanism ensures permutation invariance in neighbor orders, scalability issues arise due to the centralized system. \citet{TarMAC} adopt a similar approach, but each agent solely relies on an Attention network to determine message priorities based on the Attention weights of other agents. More recently, \citet{async_att_explore} have also leveraged Attention Networks in \ac{MARL} for real-robot exploration tasks within simplified grid worlds. 

Another line of work to facilitate agent communication is by constructing overhead 2-D maps, from where ego agents can observe information of others~\citep{forMIC, spatial_intention_map}. Inspired by social insects such as ants, \citet{forMIC} allow agents to communicate with temporary pheromone trails in 2-D maps. On the other hand, \citet{spatial_intention_map} generates agents' 2-D intention maps to inform others about their next actions. As a form of visual information, orders of neighboring agents in overhead 2-D maps are naturally permutation invariant. However, this approach restricts working environments to single-level terrains. 

In our work, we use a lightweight encoder architecture inspired by PointNet~\citep{point_net}. This architecture allows each agent to process information from a variable number of permutation invariant entities, effectively facilitating intelligent robot behaviors that dynamically focus on key entities during policy execution.

\begin{figure*}[!t]
    \centering
    \includegraphics[width=\textwidth]{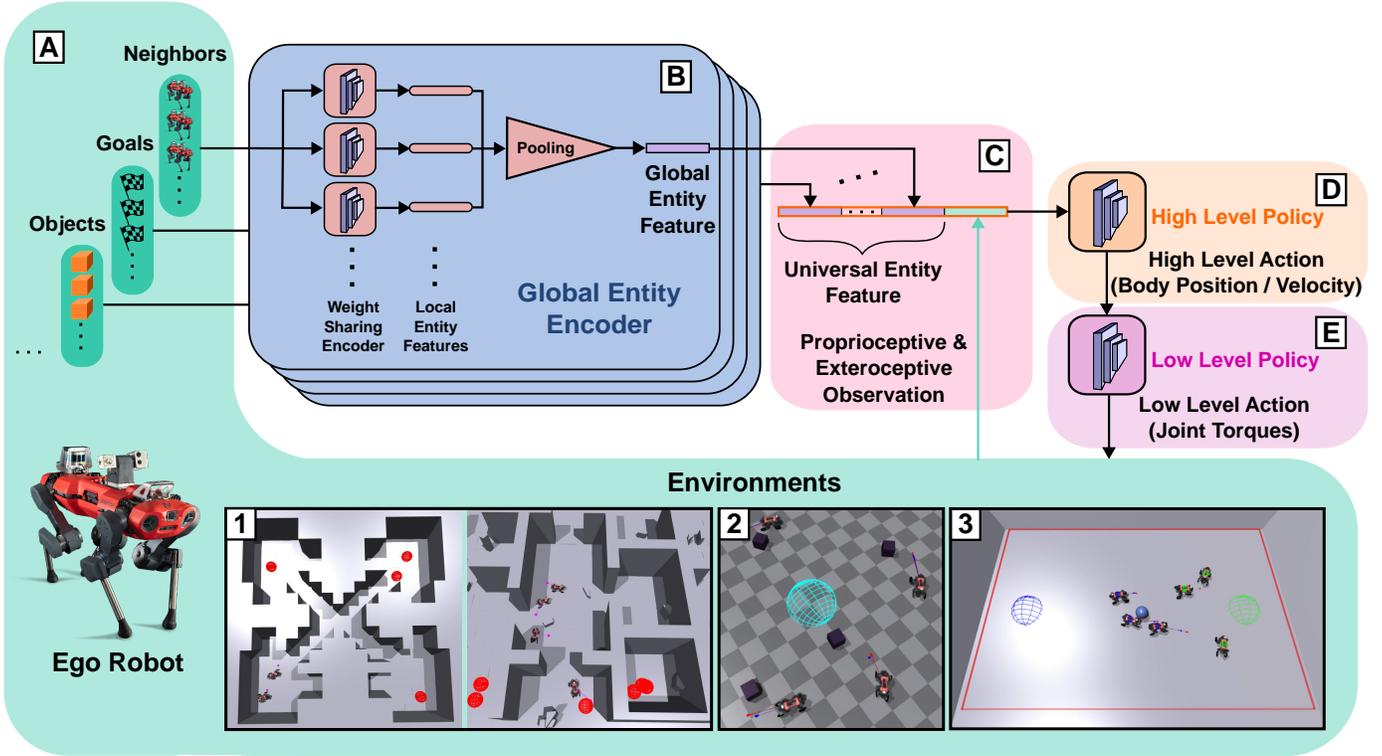}
    \caption{Pipeline overview for a decentralized ego robot. (A) The ego robot and its environments: robot can observe entities such as neighbors, goals, objects, etc. (A-1) \ac{MRMG} navigation environments. (A-2) Box packing environment. (A-3) Soccer environment. (B) All entities belonging to one category are passed to one \ac{GEE}. Their entity states are first passed into individual weight sharing encoders to get local entity features. The local entity features are then max-pooled to obtain the global entity feature for this entity category. (C) All global entity features belonging to different entity categories are concatenated to form the universal entity feature, which is then concatenated with the ego robot's proprioceptive \& exteroceptive observations from the environment to form the input to the high-level policy. (D) The high-level policy is an \ac{RL} policy network that outputs high-level actions for the ego robot. The high-level actions can be the target body position or body velocity, depending on required task settings. (E) The low-level policy takes high-level actions and outputs joint torques to control the ego robot. The low-level policy is pretrained based on the work by~\citet{lee2022control} and fixed when training high-level policies.}
    \label{fig:overview}
\end{figure*}

\section{Method}
\label{sec:methods}

\subsection{Overview}
The overview of our \ac{MARL} pipeline is shown in \cref{fig:overview}. 
We adopt a hierarchical architecture, similar to~\citep{sim2real_legged_robots}, where high-level policies produce body position or velocity commands to the low-level policies.
In this work, our focus lies on training high-level policies. 
The low-level policy is pre-trained and fixed. 
Each agent observes entities using separate \ac{GEE}s per entity type. (\cref{fig:overview}A, B)
These \ac{GEE}s produce fixed size feature vectors via pooling operations for each entity type (\cref{fig:overview}C).
These features, along with the robot's local observations, are then concatenated and provided to the downstream part of the network. The local observations include both proprioceptive and exteroceptive measurements.
The \ac{GEE} is trained end-to-end via \ac{RL}.

For our hardware experiments, we establish a robot communication pipeline using the gRPC framework~\cite{grpc}, as described in \cref{app:grpc}.

\subsection{Problem Formulation}
\label{sec:problem_formulation}
We model the multi-entity problems mentioned in \cref{fig:opener} as \ac{Dec-POMDP}s. A  \ac{Dec-POMDP} is defined by a tuple:
\begin{align*}
    <\mathcal{A}, \mathcal{S}, \mathcal{U}, \Omega, \mathpzc{O}, \mathpzc{P}, \mathpzc{R}, \gamma> \,,
\end{align*}
where $\mathcal{A}$ denotes the set of all robots $\mathcal{A} = \{a^n\}^N_{n=1}$, and $\boldsymbol{s} \in \mathcal{S}$ is the global state of the environment. Unlike previous works~\citep{legged_payload, sim2real_legged_robots, soccer_real}, where the number of robots $N$ is fixed for all environments, we allow $N$ to be a random system state. Since we use identical robots, they all share the same action space $\mathcal{U}$ and observation space $\Omega$. We use $\boldsymbol{o} \in \Omega^N$ and $\boldsymbol{u} \in \mathcal{U}^N$ to represent the joint observations and joint actions of all robots, where $\boldsymbol{o}=(o^1, \dots, o^N)$ and $\boldsymbol{u}=(u^1, \dots, u^N)$. Due to partial observability, each robot $a^n$ cannot observe the full global state $\boldsymbol{s}_t$ at time step $t$, but a partial observation state $o_t^n \in \Omega$ output by the observation function $\mathpzc{O}(\boldsymbol{s}_t, a^n)$. $\mathpzc{P}$ denotes the transition dynamics, with $\mathpzc{P}(\boldsymbol{s}_{t+1}|\boldsymbol{s}_t, \boldsymbol{u}_t)$ being the probability of transitioning to the next global state $\mathbf{s}_{t+1}$ by taking the joint action $\boldsymbol{u}$ at global state $\mathbf{s}_t$. $\mathpzc{R} \colon \mathcal{S} \times \mathcal{U} \rightarrow \mathbb{R}$ is the common reward function shared by all robots, and $\gamma \in [0, 1)$ is the discount factor. The robots collectively roll out a trajectory $\boldsymbol{\tau} = (\mathbf{s}_0, \boldsymbol{u}_0, \mathbf{s}_1, \dots)$, based on the shared decentralized policy $\boldsymbol{\pi}(u^n|o^n): \Omega \times \mathcal{U} \rightarrow [0, 1]$. The goal of all robots is to find the best policy to maximize the cumulative discounted reward $\displaystyle \mathop{\mathbb{E}}_{\boldsymbol{\tau} \sim \boldsymbol{\pi}} \left[\mathop{\sum}_{t=0}^{\infty} \gamma^t \mathpzc{R}(\mathbf{s}_t, \boldsymbol{u}_t) \right]$.

We used two different action spaces. 
For \ac{MRMG} navigation, the high-level policy commands target body $xy$ positions in the current baseframe. This is then tracked by a position tracking controller by \cite{tranzatto2022cerberus} for robot experiments.
For other problems, we used target body velocities consisting of linear $xy$ velocity and yaw rates in base frame for simplicity.
More implementation details can be found in \cref{app:hierarchy}, and we refer the interested reader to~\cite{lee2022control} for more details about the low-level policy.
To ensure that the outputs of our high-level policies adhere to the constraints of our robot's physical capabilities, we adopt a bounded action space based on Beta distribution~\citep{beta, beta_ppo}. We refer readers to \cref{app:beta} for more information. 

The rewards and observations for our three \ac{Dec-POMDP}s are displayed in \cref{tab:reward} and \ref{tab:observation} in the Appendix, for brevity.

In all experiments, the agents do not have access to the global map information. Each agent only observes a \SI{4.35}{\meter}$\times$\SI{2.85}{\meter} rectangular local height map around itself.

\subsection{Global Entity Encoder}
\label{sec:gee}
Since we let the number of robots $N$ be random, each robot $a^n$ needs to observe flexible numbers of neighbors in a fixed size observation vector $o^n$. Such variability of numbers also applies to other task-related entities, such as navigation goals, boxes, opponents, etc. To this end, we approximate the observation function as:
\begin{equation}
    \label{eqn:obs_func}
    o^n = \mathpzc{O}(\boldsymbol{s}, a^n)  \simeq  \mathpzc{O}(\mathpzc{f}(\boldsymbol{s}, a^n), s^n) \,,
\end{equation}
where $\mathpzc{f}(\boldsymbol{s}, a^n)$ is the universal entity feature (\cref{fig:overview}C) and $s^n$ is the local observation of robot $a^n$. 
The universal entity feature is the concatenation of a number $E$ of global entity features, each corresponding to a unique entity category in the environment:
\begin{equation}
    \label{eqn:universal_entity_feat}
    \mathpzc{f}(\boldsymbol{s}, a^n) = [\mathpzc{g}^1(\boldsymbol{s}, a^n), \dots, \mathpzc{g}^E(\boldsymbol{s}, a^n)] \,.
\end{equation}
Each $\mathpzc{g}$ represents a global entity feature from a \ac{GEE}. In our box packing task, for example, two \ac{GEE}s are used ($E = 2$), where one is applied to all neighbors while another is applied to all boxes. Depending on the number of entity categories in the environment, any number of \ac{GEE}s can be chosen for a given task.

The operation to get a global entity feature $\mathpzc{g}$ and the universal entity feature $\mathpzc{f}$ is shown in \cref{fig:overview}B and \ref{fig:overview}C. Similarly to PointNet~\citep{point_net}, for each robot $a^n$, all observed entities from the same category are passed to a weight sharing MLP to get corresponding local entity features; then, all local entity features are max-pooled to obtain an associated global entity feature $\mathpzc{g}$. All global entity features representing different entity categories are concatenated to get the universal entity feature $\mathpzc{f}$. Together with the robot's proprioceptive and exteroceptive observations from the environment, the observation state $o^n$ is passed to the Policy Net to get a single robot's action $u^n$. 

\subsection{Training Environments}
\label{sec:training_envs}

Here we describe the training environments for the three multi-entitiy problems. All environments have time limit of \SI{45}{\second} and the number of entities is randomized per episode.

\subsubsection{\ac{MRMG} Navigation}
\label{sec:training_mrmg}
The \ac{MRMG} navigation task requires a group of robots to visit multiple goal positions. Each goal has to be visited at least once by any robots. 
We represent each goal as a \SI{0.5}{\meter} radius sphere and initialize it \qtyrange{15}{30}{\meter} away from the centroid of the robot group. 
An episode ends successfully when all goals are reached, or ends in failure if the time limit is exceeded. 
We design different terrains as shown in \cref{fig:overview}A-1.
During training, each episode involve a maximum of four robots and up to four goals, with the number of entities varying between episodes. Note that in policy deployment, the number of entities can be higher than the maximum values in the training stage.

A terrain curriculum similar to \citet{walk_in_min} is used, where the terrain becomes more difficult if a task is completed successfully, or becomes easier if no goal is reached. 
Terrain difficulty is associated with certain attributes, such as the narrowness of corridors, ground flatness, height of stairs, etc.

\subsubsection{Box Packing}
\label{sec:training_packing}
The box packing task requires a group of robots to push a group of boxes to a packing site, as shown in \cref{fig:overview}A-2. 
The packing site is a \SI{1.5}{\meter} radius sphere, while the robots and boxes are randomly positioned in various direction within \qtyrange{1.5}{10}{\meter}. 
An episode ends in failure if any robot falls or if the time limit is exceeded.
Similar to \ac{MRMG} navigation, each training episode involve a maximum of four robots and up to four boxes, with the number of entities varying between episodes.

We uniformly sample the initial distances of the robots and boxes from the packing site. 
if the robots successfully push all boxes to the packing site, we increase the upper bounds of the sampling range by \SI{1}{\meter}. 
Otherwise, we decrease both upper bounds by \SI{1}{\meter}. 
The sampling range is capped between \qtyrange{2}{10}{\meter} for robots and \qtyrange{1.5}{10}{\meter} for boxes.

\subsubsection{Soccer}
\label{sec:training_soccer}
Two groups of robots try to push a ball into the other team's goal while defending their own. 
As shown in \cref{fig:overview}A-3, the goal is a \SI{1}{\meter}-radius sphere located near the end of each side of the soccer environment. The field is a \SI{18}{\meter}$\times$\SI{12}{\meter} rectangle and \SI{45}{\degree} slopes are installed at the field boundaries to prevent the ball from rolling away. 
At the beginning, all robots are lined up in a row in front of their respective goals, and the soccer ball is initialized at the center with random initial velocity.
An episode terminates whenever the ball enters either team's goal, or if the time limit is exceeded.
We assign 1--3 robots on the training side, while 0--3 robots are on the opponent side. 

We implemented a self-play curriculum, a popular method in competitive \ac{RL} implementations~\citep{hide_seek, soccer_sim, soccer_real, multi_agent_competition}. 
 In this strategy, the robots on the opponent team are equipped with earlier policies used against the training team. This allows both teams' strategies to improve simultaneously.

\begin{figure*}
    \centering
    \includegraphics[width=\textwidth]{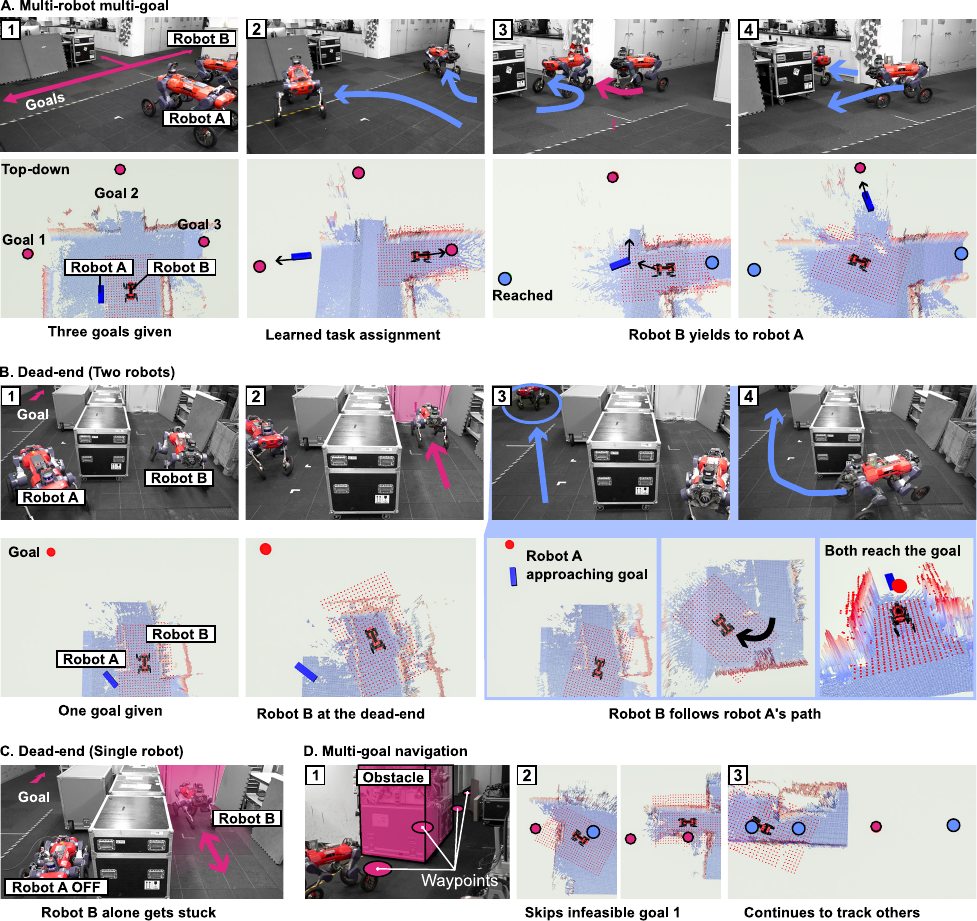}
    \caption{Robot experiments. (A) \ac{MRMG} navigation with two robots and three goals. (B, C) Dead-end experiment. (D) Single-robot Multi-goal navigation.  }
    \label{fig:robot_exp}
\end{figure*}

\begin{figure*}[!t]
    \centering
    \includegraphics[width=\textwidth]{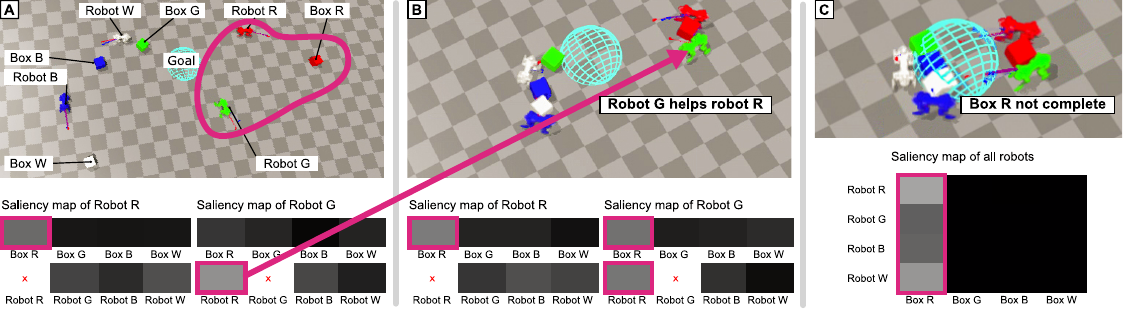}
    \caption{
    Saliency maps of two robots during the box packing task.
   The four robots have to move the boxes into the goal point.
    (A) At the beginning, robot G does not focus on any box but attends to the states of robot R. (B) Then robot G joins robot R to transport the box R together. 
    (C) Once all boxes are at the goal except for box R, all robots shift their focus to box R.}
    \label{fig:saliency}
\end{figure*}

\section{Experiments and Results}

We validate our approach in a series of hardware and simulation experiments.
Our tests on real robots, as discussed in \cref{sec:real_exp}, qualitatively showcase the effectiveness of our policies. \cref{sec:analysis_collaboration} quantitatively analyzes our policies' internal reasoning during intensive robot collaborations and examines the impact on task performance when varying the number of collaborators.
We further conduct ablation studies in \cref{sec:ablation} to show the contribution of the \ac{GEE} architecture and communication on robots' intelligent collaborative behaviors.
Finally, we compare our \ac{MARL} policy with a centralized optimization-based controller in a simplified multi-agent task in \cref{sec:existing_method} and show that our policy automatically learns near-optimal solutions.

\subsection{Robot Navigation}
\label{sec:real_exp}
We conduct three different multi-entity navigation experiments using two wheeled-legged quadrupedal robots~\cite{bjelonic2019keep}.
The results are shown in the supplementary video.
In each experiment, the same control policy is used across all robots, and the robots' states are shared through gRPC communication---see \cref{app:grpc}.
Robots in all experiments do not possess the global map or share any map information. Instead, each robot only observes a \SI{4.35}{\meter}$\times$\SI{2.85}{\meter} rectangular local height scan centered on itself, as indicated by the red points in the top-down views in \cref{fig:robot_exp}.

\subsubsection{\ac{MRMG} Navigation}
Two robots are tasked with reaching three shared goals, as shown in \cref{fig:robot_exp}A-1. The task is completed once all goals are reached, regardless of which robot reaches each goal. 
Each robot strategically chooses the closest goal while leaving the farther goal to the other robot (\cref{fig:robot_exp}A-2). 
Upon finishing the first two goals (goal 1 and goal 3 in the figure), both robots try to reach goal 2 and meet at the narrow corridor. 
At the entrance, the robots avoid collision, and one robot stops and leaves the last goal to the other robot.

The distribution of goals and collision avoidance behavior observed in this experiment is learned, with no handcrafted state machine or task scheduling as seen in~\citep{mamg_pick_up_delivery, multi-obj-search}.

\subsubsection{Dead End}
In the experiment shown in \cref{fig:robot_exp}B and C, two robots are given a single waypoint to reach. 
Unlike the previous experiment, both robots have to reach the waypoint.
Robot A has an obstacle-free path to the goal in the line of sight, whereas robot B starts from a corridor with a dead end. 
It is important to note that neither robot has access to a global map; they rely solely on local perception.

At the beginning, both robots navigate towards the goal. Then robot B encounters the dead end. 
Observing the robot A's successful path to the goal, the trapped robot B adapts and follows robot A's path, successfully escaping from the dead end. 
In contrast, \cref{fig:robot_exp}C shows a single robot scenario. Without its collaborator, robot B fails to find the exit and becomes trapped in a local minimum, oscillating back and forth.

This experiment demonstrates the effectiveness of our approach: the control policy observes the other agents' behavior and utilizes this information to resolve navigation challenges.

\subsubsection{Multi-Goal Navigation with Obstacles}

Mobile robots are often tasked with multiple waypoints, e.g., path-following or multi-goal delivery tasks.
Typically, the goals are given to the robot in a fixed order~\citep{smug_nav, bjelonic2019keep, multi-obj-search}.
Additional state machines or human intervention may be necessary if a given waypoint becomes unreachable due to unexpected obstacles or environmental changes. 

Our approach addresses this challenge by letting the policy determine the visiting order of waypoints dynamically.
In another experiment, we give a robot multiple intermediate waypoints along a straight path, one of which is blocked by obstacles---see \cref{fig:robot_exp}D.
The robot navigates to the waypoints sequentially, automatically skipping the blocked waypoint. As shown in \cref{fig:robot_exp}D-2, our multi-entity RL method can delegate such high-level decision-making challenges to the policy.

\subsection{Learned Collaborative Behaviors}
\label{sec:analysis_collaboration}
We then analyse the emergent collaborative behaviors with two experiments on box packing task. For both experiments, we use the same policy trained in the environment depicted in \cref{fig:overview}A-2 following the training procedure in \cref{sec:training_packing}.

\begin{figure}[!t]
    \centering
    \includegraphics[width=\columnwidth]{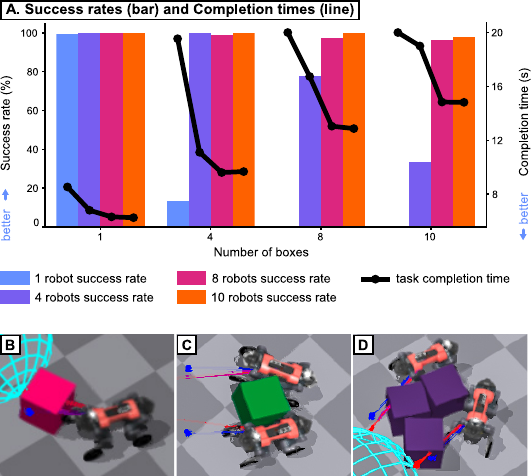}
    \caption{Impact of Robot Numbers on Packing tasks. (A) Success Rate (bar columns) and Completion Time (dotted lines) of 1-10 robots packing 1-10 boxes. (B) 1 robot transports a box by walking sideways. (C) 2 robots are faster by squeezing the box in between them. (D) 2 robots can transport more than 1 box at the same time.}
    \label{fig:number_impact_packing}
\end{figure}

\subsubsection{Dynamic Focus on Neighboring Entities}
We conduct a sensitivity analysis to examine what entities each agent focuses on at each time step.
Similarly to \cite{Lee_2020}, we compute the gradient magnitude of the output feature of \ac{GEE} with respect to each entity state.
Intuitively, if an entity has no impact on the robot's behavior, perturbing the entity would not change the output of the \ac{GEE}.

The resulting saliency map during the box packing task is shown in \cref{fig:saliency}. 
It is observed that the robots strategically allocate their attention to boxes and adjacent robots that they may interact with. This focus dynamically adapts in real-time to the evolving situation. 
This dynamic entity prioritization is a learned behavior, a significant advancement over traditional approaches. 
Conventional methods using heuristic-based task assignments lack the flexibility to adjust to changing contexts in this manner.


\subsubsection{Adaptation to Higher Entity Numbers}
We examine whether an increase in the number of robots leads to improved task performance, as well as the effectiveness of our policies across different numbers of targets.
We trained a policy for the box packing task with a randomized number of robots and boxes, ranging up to four.
We then execute the same policy with different combinations of robots and boxes.

\cref{fig:number_impact_packing}A shows the success rate and completion time of different numbers of robots packing boxes within a \SI{20}{\second} time limit. We observe that, as the number of robots increases, the task is more likely to succeed and requires less time, especially when for higher numbers of boxes.

It is noteworthy that the completion time decreases as more robots are involved, even in scenarios with just one box. This tendency persists across all combinations of robot and box numbers once the number of robots surpasses the number of boxes.

The decrease in completion time would not be as significant if each box were always assigned to a single robot. We observe that when one robot pushes one box, it uses its side to avoid the box from slipping away (\cref{fig:number_impact_packing}B). The presence of more robots allows for faster transportation by squeezing boxes between them, as shown in \cref{fig:number_impact_packing}C and \cref{fig:number_impact_packing}D.

This example shows a benefit of learning collaborative behaviors through RL. Such behavior cannot be easily handcrafted, highlighting the effectiveness of our approach in solving complex problems.

\subsection{Enhanced Coordination with GEE}
\label{sec:ablation}
We identified two key components crucial to enhancing multi-robot coordination in our experiments: the observation of neighboring agents (neighbor awareness) and the Global Entity Encoder (\ac{GEE}). 

We conduct two ablation studies focusing on their roles in addressing multi-entity problems.
We conducted two experiments with \ac{MRMG} navigation (see \cref{fig:overview}A-1) and soccer environments (see \cref{fig:overview}A-3). The training procedures of the two policies are described in \cref{sec:training_mrmg} and \cref{sec:training_soccer}.

\subsubsection{Neighbor Awareness}

\begin{figure}[!t]
    \centering
    \includegraphics[width=\columnwidth]{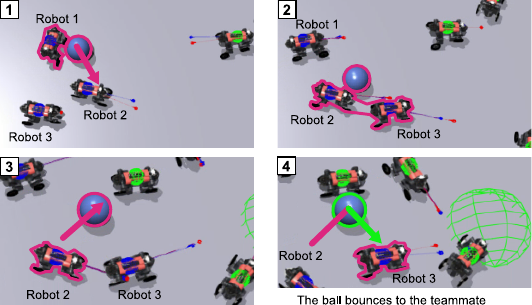}
    \caption{Ball passing maneuver during the soccer game. The robots 1, 2, 3 advance towards the right side.}
    \label{fig:full_vs_no_comm}
\end{figure}

We first assess the influence of neighbor awareness on a 3 vs. 3 soccer game, as depicted in \cref{fig:full_vs_no_comm}.
Two teams are set to compete: one with full awareness of both teammates and opponents (neighbor-aware), and the other devoid of any awareness regarding either teammates or opponents (neighbor-unaware). Each policy for each team is trained independently with its corresponding setup.

The neighbor-aware team wins 86.4\% of the matches against the neighbor-unaware team.
In the snapshots in \cref{fig:full_vs_no_comm}, we observe that robots in the fully aware team have learned collaborative strategies such as spreading out and passing balls to teammates (1 \& 2), sometimes passing balls to their opponents and catch the ball when it bounces back (3 \& 4).
The neighbor-unaware team often collide with their teammates and does not exhibit such collaboration.

\subsubsection{Global Entity Encoder}

\begin{figure}[!t]
\centering
{\includegraphics[width=0.479\textwidth]{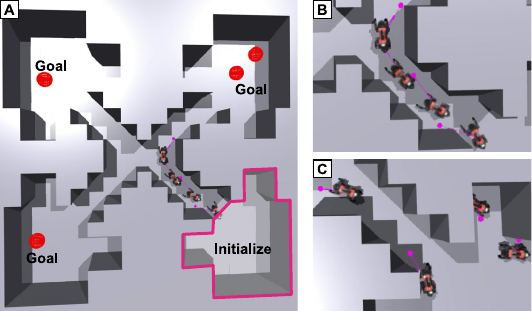}}

\caption{Ablation study on \ac{GEE}. (A) \ac{MRMG} navigation test environment. Four robots are spawned in the bottom right corner, and the goals are sampled from the other three corners. (B) With GEE, when one robot finds a way out of the local minima, robots exhibit the leader-following behavior. (C) When GEE is replaced with a naive concatenation of features, the leader-following behavior does not emerge.}
\label{fig:ablation}
\end{figure}

We now assess the role of \ac{GEE} in enhancing the collaborative task performance.
We replace the \ac{GEE} with naively concatenating entity features and compared the success rate for \ac{MRMG} navigation task (see \cref{fig:overview}A-1).
Training the concatenation model is the same as training the GEE models as described in \cref{sec:training_mrmg}. 
The only difference is that we have to set a maximum entity number for each entity, and pad the concatenated features with zeros if the environment has fewer entities than the max number. We set the maximum entity numbers as those used when training the GEE model (4 robots, 4 goals). 
Note that the concatenation model cannot handle more than the max entity numbers but the GEE models can.

The GEE model yields a success rate of 96.7\% on the task and an average completion time of \SI{20.3}{\second}, compared to 86.4\% and \SI{22.7}{\second} for the concatenation. 
\cref{fig:ablation} shows their different behaviors in one example.
In the cross-like corridor environment, the policy with GEE exhibits the leader-following behavior in \cref{fig:ablation}B. The behavior is also shown in \cref{fig:robot_exp}B. Once one of the robots finds a way out of the first room, the others follow the robot.

Our results show that the \ac{GEE} significantly improves the performance of neighbor-aware policies by effectively leveraging global contextual information.

\subsection{Comparison to an Optimal Control Approach}
\label{sec:existing_method}

\begin{figure}[!t]
    \centering
    \includegraphics[width=0.95\columnwidth]{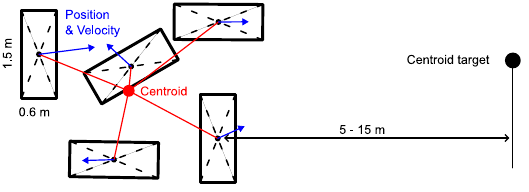}
    \caption{Two-dimensional centroid control problem. A controller (Ours or MPC) controls the position and velocity of cuboids to let their centroid (red dot) track the target position. The cuboids are randomly spawned between \qtyrange{5}{15}{\meter} away from the goal.}
    \label{fig:centroid}
\end{figure}

To evaluate the optimality of our learned control policy, we conduct a comparative analysis with a centralized \ac{MPC} approach for the centroid control task illustrated in \cref{fig:centroid}.
In this task, the objective is to generate position and velocity targets for each cuboid to move the group's centroid to a goal position while avoiding collisions.

Our policy is trained with 1--6 cuboid agents in each environment and benchmarked against a simplified version of the planner by \citet{flavio_mpc} as the baseline.
The control task is formulated as a kinematic planar system involving $R$ cuboids.
Both approaches solve for the desired velocity and position of each cuboid.

\subsubsection{MPC Formulation}
\label{sec:mpc_formulation}
We approximate each robot $i$ in a squad of $R$ robots as a cuboid with a two-dimensional position $\mathbf{p}_i \in \mathbb{R}^2$ and a yaw angle $\psi_i \in \mathbb{R}$. Given a discrete time horizon $N \in \mathbb{N}$ and a step size $\Delta t > 0$, and having defined the $i$th robot's state and input sampled at time step $k$ as $\mathbf{x}_{i, k} := [\mathbf{p}_{i, k}^\top \; \psi_{i, k} \; \dot{\mathbf{p}}_{i, k}^\top \; \dot{\psi}_{i, k}]^\top$ and $\mathbf{u}_{i, k} := [\ddot{\mathbf{p}}_{i, k}^\top \; \ddot{\psi}_{i, k}]^\top$, respectively, we formulate our optimal control problem (OCP) as the following nonlinear programming problem:
\begin{subequations}\label{eqn:kcl_ocp}
\begin{align}
    \min_{\mathbf{X}, \mathbf{U}}\;
    &\sum_{k = 0}^{N - 1} \Bigg[ \| {\sum_i \mathbf{p}_{i, k}}/{R} - \mathbf{c}^\ast \|_{1, \mathbf{W}_{c}} + \nonumber \\
    &\qquad\;\;\, \| {\sum_i \dot{\mathbf{p}}_{i, k}}/{R} - \dot{\mathbf{c}}^\ast \|_{\mathbf{W}_{\dot{c}}}^2 + \nonumber \\
    &\qquad\;\; \sum_{i = 0}^{R - 1} \| \mathbf{x}_{i, k} - \mathbf{x}_{i, k}^\ast \|_{\mathbf{W}_x}^2 + \| \mathbf{u}_{i, k} \|_{\mathbf{W}_u}^2 \Bigg] \label{eq:kcl_objective} \\
    \text{s.t.}\quad &\mathbf{x}_{i, 0} = \mathbf{x}_{i, m}\,, \forall i \,, \nonumber \\
    &\mathbf{x}_{i, k+1} = \mathbf{A}\mathbf{x}_k + \mathbf{B} \mathbf{u}_k \,, \forall i \,, \forall k \,, \nonumber \\
    &\| \mathbf{p}_{i, k} - \mathbf{p}_{j, k} \| \geq 2\rho \,, \forall i \,, \forall j \,, \forall k \,, \nonumber \\
    &| {}_b\dot{\mathbf{p}}_{i, k}^x | \leq v_\text{max}^x \,, \forall i \,, \forall k \,, \nonumber \\
    &| {}_b\dot{\mathbf{p}}_{i, k}^y | \leq v_\text{max}^y \,, \forall i \,, \forall k \,, \nonumber \\
    &| \dot{\psi}_{i, k} | \leq \omega_\text{max} \,, \forall i \,, \forall k \,, \nonumber
\end{align}
\end{subequations}
where $\mathbf{c}^\ast$ and $\dot{\mathbf{c}}^\ast$ are the reference centroid position and velocity, respectively, while $\mathbf{x}_{i, k}^\ast$ is the reference state of the $i$th robot at time step $k$. $\mathbf{W}_\square$ is a positive definite diagonal matrix for the weighted $1$- and $2$-norms $\| \cdot \|_{1, \mathbf{W}_\square}$ and $\| \cdot \|_{\mathbf{W}_\square}$, respectively; $\mathbf{x}_{i, m}$ is the $i$th robot's measured state. We integrate the input accelerations using a semi-implicit Euler method, which translates to the following system matrices:
\begin{equation*}
\mathbf{A} := \begin{bmatrix}
\mathbf{I}_3 & \Delta t\mathbf{I}_3 \\
\mathbf{0} & \mathbf{I}_3
\end{bmatrix} \,, \;
\mathbf{B} := \begin{bmatrix}
\Delta t^2\mathbf{I}_3 \\
\Delta t\mathbf{I}_3
\end{bmatrix} \,.
\end{equation*}
We provide more details about our \ac{MPC} implementation in \cref{app:mpc}.

\subsubsection{Results}

\begin{table*}
    \centering
    \caption{Comparison in centroid control task.}
    \label{tab:centroid}
    \begin{tabular}{>{\centering\arraybackslash}m{3.0cm}|>{\centering\arraybackslash}m{3.0cm}|>{\centering\arraybackslash}m{3.0cm}|>{\centering\arraybackslash}m{3.3cm}|>{\centering\arraybackslash}m{3.0cm}}
        \hline
        & Tracking error (m) & Settling time to 0.1 m (s) & Centroid travel distance (m)  & Solver time (ms) \\
        \hline
        \textbf{Ours 4 robots} & 0.0168 $\pm$ 0.0085 & 4.70 $\pm$ 1.48 & 10.11 $\pm$ 3.03 & 0.51 $\pm$ 0.03 \\
        \hline
        \textbf{MPC~\citep{flavio_mpc} 4 robots} & 0.0070 $\pm$ 0.0030 & 5.35 $\pm$ 0.92 & 10.38 $\pm$ 1.90 & 9.61 $\pm$ 0.29 \\
        \hline
        \hline
        \textbf{Ours 10 robots} & 0.0478 $\pm$ 0.0249 & 8.02 $\pm$ 0.78 & 10.53 $\pm$ 1.82 & 0.77 $\pm$ 0.07 \\
        \hline
        \textbf{MPC~\citep{flavio_mpc} 10 robots} & 0.0044 $\pm$ 0.0020 & 5.90 $\pm$ 1.35 & 11.18 $\pm$ 2.44 & 91.03 $\pm$ 5.52 \\
        \hline
    \end{tabular}
\end{table*}

\cref{tab:centroid} shows the mean and standard deviation of three evaluation metrics for both our decentralized Multi-Entity \ac{RL} policy and the centralized MPC planner on two setups: centroid control with 4 cuboids and 10 cuboids.
We run 10 experiments with cuboids randomly positioned \qtyrange{5}{15}{\meter} away from the goal and compare the final tracking errors, settling times, centroid travel distances, and solver time of both methods on both setups.
We remark that the 10-cuboid scenario is not seen during our policy training stage, as the training environments only consist of 1 to 6 cuboids.

Both approaches exhibit a very similar performance. With 4 cuboids, \ac{MPC} leads by a narrow margin on the tracking error (\SI{\sim 1}{\centi\meter}), but it has slightly longer settling times. In all setups, our \ac{RL} policy showcases shorter centroid travel distances, suggesting more efficient goal-pursuit trajectories. The compared metrics indicate that our Multi-Entity \ac{RL} policy can find near-optimal solutions in the considered scenario.

Our decentralized policy generalizes to unseen tasks involving 10 cuboids with minimal deterioration from the 4-cuboid setup, with \SI{\sim 3}{\centi\meter} higher tracking errors and \SI{\sim 3.3}{\second} longer settling times.
The increased solver times of the \ac{MPC} approach highlight the limited scalability of conventional centralized methods. In contrast, our decentralized \ac{RL} policy maintains nearly constant computation times regardless of the number of entities involved.


\section{Conclusion}
We present a \ac{MARL} framework using permutation invariant neural network encoders capable of processing various categories of entities in a wide range of robotic environments. 
We validated our approach by implementing and testing a complete system to control \ac{MAS} with wheeled-legged quadrupeds. The system consists of decentralized multi-entity control policy and communication among robots using gRPC~\citep{grpc}.

Our real-world \ac{MRMG} navigation experiments with two robots demonstrated collaborative multi-goal navigation behaviors learned to optimize task performance without the use of heuristics.
We observed intelligent sub-goal selection, navigation, and collision avoidance.

Furthermore, we conducted controlled experiments in simulation on three distinct multi-entity tasks, aiming to showcase the scalability and generalizability of our approach.
Our experiments showed that the \ac{GEE}s enables agents to process a flexible number of entities in execution time. \ac{GEE}s enable each robot to collaborate with flexible numbers of teammates and prioritize suitable objects or goals. 
This resulted in improved overall performance compared to teams without the permutation invariant encoder structure or communication among robots.
Additionally, we conducted a sensitivity analysis on \ac{GEE}s, providing an evidence that our policy automatically adjusts its focus to different entities in the environment at each moment.

Moreover, we compared our policy to an existing optimal-control approach on a multi-agent navigation problem and showed that our policy learns a near-optimal solution in this scenario.

We plan to extend our research to encompass heterogeneous robots, such as robots equipped with different sensors or manipulators for specialized tasks. We hope this work could inspire future research on general purpose AI and robotics.
Our goal is to advance robots beyond simple lab environments with constrained task configurations. We aim to enhance their capabilities to excel in diverse real-world environments with a wide range or entities and tasks.

\section*{Acknowledgments}
We thank Yifan Liu for the support in implementing simulation environments for \ac{MRMG} navigation training. We thank Alexander Reske and Turcan Tuna for helping with the real robot experiments. 

This project was supported by the Swiss National Science Foundation through the National Centre of Competence in Digital Fabrication (NCCR dfab) and the European Research Council (Grants 852044, 101070596, and 101121321).

{\small


\bibliographystyle{IEEEtranN}}

{\appendix

\subsection{Hierarchical Policy}
\label{app:hierarchy}
Hierarchical policies have been widely applied on complex robotic tasks~\citep{central_cable-towed_load, skill_behav_diver, sim2real_legged_robots}, as modularized policies provide easier control over desired behaviors at each task level. We use a similar way to train hierarchical policies as in~\citep{sim2real_legged_robots}. First, a locomotion policy is trained according to the work by~\citet{lee2022control}. This policy takes body velocity commands and executes joint-level control on individual robots. The locomotion policy is treated as the low-level policy and fixed when training higher level policies. In this project, both the box packing task and the soccer task use two-level policy structure: the low-level policy is the locomotion policy, and the high-level policies output body velocity commands to complete specific multi-agent tasks. The \ac{MRMG} navigation task uses three-level policy structure, where the low-level policy is the same locomotion policy. Similar to the high-level policies of the other two tasks, the mid-level policy in \ac{MRMG} navigation outputs body velocity commands to control individual robots towards a shorter goal point (\qtyrange{0}{2}{\meter} from ego robot) without observing neighboring robots, and the high-level policy outputs these shorter goal points to navigate each robot to flexible numbers of far goal points (\qtyrange{15}{30}{\meter}) in a \ac{MAS}. In fact, the low-level policy is the same for all three tasks, as the desired locomotion behaviors are the same in such low level. This is the advantage of modularized hierarchical policy structure, as some parts of the policy can be shared among different tasks, thus greatly reducing the time to train robots from scratch.

\subsection{Bounded Action Space using Beta Distribution}
\label{app:beta}
For stochastic policy gradient methods such as PPO~\citep{ppo}, a practical limitation is that the Gaussian distribution has infinite support over the action space. This creates challenges to constrain policy outputs for robots' physical limitations. For example, the maximum speed of our robots is 2 m/s in longitudinal directions, 1 m/s in lateral directions and 1.5 rad/s about vertical axis. Therefore, the body velocity commands as action outputs from our high-level policies must be within these ranges, which cannot be satisfied by Gaussian distribution. Inspired by~\citep{beta} and~\citep{beta_ppo}, we replace Gaussian distribution with Beta distribution as the stochastic action distribution function in our PPO algorithm~\citep{ppo}. The benefits of Beta distribution is that the output action distribution is bounded between 0 and 1, so it can be linearly scaled to any bounded action range without suffering the loss of sampling accuracy. The policy network outputs parameters $\alpha$ and $\beta$ for Beta action distribution:

\begin{equation}
    \label{eqn:beta_distribution}
    \textit{Beta}(x; \alpha, \beta) = \frac{1}{B(\alpha, \beta)} x^{\alpha-1} (1-x)^{\beta-1}
\end{equation}

The shape of the distribution function depends only on the two parameters $\alpha$ and $\beta$, and $B(\alpha, \beta)$ is called the beta function, which also depends on $\alpha$ and $\beta$ and ensure the total probability equals to 1.

Stochastic actions are sampled from the Beta distribution during training to encourage exploration. During policy execution, we take the distribution mean $\frac{\alpha}{\alpha + \beta}$ as the deterministic action.

\subsection{\ac{MPC} Baseline Implementation Details}
\label{app:mpc}
We implement a simplified \ac{MPC} planner based on the work by~\citet{flavio_mpc} and delineate its formulation in \cref{sec:mpc_formulation}. In this section, we lay out more implementation details for our \ac{MPC} controller to complete the multi-agent problem in \cref{sec:existing_method}. 

To prevent robot collisions, we enforce a minimum distance of $2\rho$ between all pairs of robots ($\rho > 0$). The quantities ${}_b\dot{\mathbf{p}}_{i, k}^x$ and ${}_b\dot{\mathbf{p}}_{i, k}^y$ represent the $x$- and $y$-components of the $i$th robot's linear velocity at time step $k$ expressed in its base frame. Finally, $v_\text{max}^x$ and $v_\text{max}^y$ denote the maximum forward and sideways linear velocities, while $\omega_\text{max}$ is the maximum yaw rate. We remark that we adopted a $1$-norm for the centroid tracking cost in \eqref{eq:kcl_objective} to prevent steep gradients for distant goal positions $\mathbf{c}^\ast$ that would cause numerical issues.

We implement our MPC in C++ using Ungar~\cite{DeVincenti-IROS-23} and send the optimized trajectories to our low-level PD controller through Python bindings~\cite{pybind11}. Specifically, the output planar linear velocities and yaw rates are converted into force and torque signals to drive each individual cuboid in the simulator. We use the parameters listed in \cref{tab:mpc_params} for all our experiments.

\begin{table}
    \centering
    \caption{MPC Parameters}
    \label{tab:mpc_params}
    \begin{tabular}{c|c}
    \toprule
        $N$ & $30$ \\
        $\Delta t$ & $0.0\overline{3}$ \\
        $\rho$ & $0.8$ \\
        $\mathbf{W}_c$ & $\mathrm{diag}(10, 10)$ \\
        $\mathbf{W}_{\dot{c}}$ & $\mathrm{diag}(0.6, 0.6)$ \\
        $\mathbf{W}_{x}$ & $10^{-2}\mathrm{diag}(0, 0, 1000, 4, 4, 4)$ \\
        $\mathbf{W}_{u}$ & $10^{-5}\mathrm{diag}(1, 1, 1000)$ \\
        $v_\text{max}^x$ & $2$ \\
        $v_\text{max}^y$ & $1$ \\
        $\omega_\text{max}$ & $1.5$ \\
    \bottomrule
    \end{tabular}
\end{table}

\begin{figure}
    \centering
    \includegraphics[width=\columnwidth]{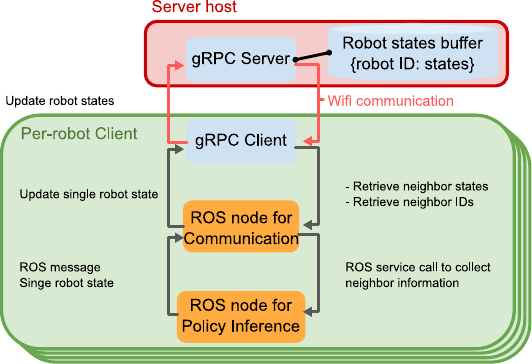}
    \caption{gRPC communication framework}
    \label{fig:gRPC}
\end{figure}

\subsection{gRPC Communication framework}
\label{app:grpc}
We implemented a multi-robot communication framework in ROS using the gRPC framework~\citep{grpc}. gRPC uses Protocol Buffers~\citep{protobuf} to serialize messages between a server and its clients distributed in different machines. In our case, the server collects the most recent messages from all clients periodically and sends them to any client that makes the request. Each robot initiates a client at the beginning, which constantly sends out its own message to the server and requests other robots' messages when needed by the policy net. The server-client communication framework is illustrated in \cref{fig:gRPC}. A gRPC server contains a message buffer with robot ID-message pairs. A robot ID is a unique number for each robot and a message is the agent state information passed to the \ac{GEE}. Every gRPC client subscribes to the server host at the beginning and creates gRPC Stubs to call service methods for message exchange with other clients. Inside each robot, a gRPC client is instantiated by the robot's ROS communication node, which is also responsible for transferring messages between the gRPC client and the navigation planner through standard ROS communication protocol.

\begin{table}[t!]
    \centering
    \caption{Rewards}
    \label{tab:reward}
    \begin{tabular}{l l l}
      \multicolumn{3}{c}{\ac{MRMG} Navigation} \\
    \hline
         Reward & Description & Scale \\ \hline
        Termination & +1.0 if game success & 10.0 \\
        Distance to Goal & + $exp(- (\text{distance to unreached goals})^2)$ & 5.0 \\
        Motion bonus & + clip(\text{moving speed}, 0.0, 1.0) & 1.0 \\
        Neighbor distance & -1.0 if too close (\textless 1 m) & 1.0 \\
        Collision & -1.0 if robot collides & 2.0 \\
  \hline
    \\ 
  \multicolumn{3}{c}{Box Packing} \\
    \hline
         Reward & Description & Scale \\ \hline
        Termination & +1.0 if game success & 5.0 \\
        Progress & +1.0 ($N_{\text{completed boxes}} / N_{\text{total box count}}$) & 0.25 \\
        Box velocity & +1.0 $\sum_{\text{boxes}} ( v_{\text{box}} \cdot \text{direction to goal})$& 0.1 \\
        Box position & +1.0 $\sum_{\text{boxes}} exp( - ( p_{\text{box}} -p_{\text{goal}})^2)$& 0.5 \\
        Neighbor distance & -1.0 if too close (\textless 1 m) & 1.0 \\
  \hline
    \\ 
  \multicolumn{3}{c}{Soccer} \\
    \hline
         Reward & Description & Scale \\ \hline
        Termination & +1.0 if wins, -1.0 if loses & 5.0 \\
        Ball velocity & +1.0 $( v_{\text{ball}} \cdot \text{direction to goal})$& 1.0 \\
        Ball position & +1.0 $\exp( - ( p_{\text{ball}} - \text{goal position})^2)$& 0.25 \\
        Neighbor distance & -1.0 if too close (\textless 1 m) & 1.0 \\
  \hline
    \end{tabular}
\end{table}

The gRPC communication pipeline is simple to use in real-world applications and can connect a flexible number of robots. When a new robot is activated and subscribed to the server host, its unique ID is saved in the message buffer in the server. When a robot requires communication with other robots, it can request other robots' messages with their IDs. After the robot executes its action and arrives at a new state, its message gets automatically updated in the message buffer.

It is worth noting that the server host can be any computer with the same internet connection as the clients, including any one of the clients. In our real-world experiments, we use a separated computer as the server host.


\begin{table*}[!t]
    \centering
    \caption{Robot Observation in all policies}
    \label{tab:observation}
    \begin{tabular}{>{\raggedright\arraybackslash}p{4cm} >{\raggedright\arraybackslash}p{4cm} >{\raggedright\arraybackslash}p{6cm} >{\raggedright\arraybackslash}p{2cm}}
        \specialrule{.15em}{.05em}{.05em}
        Policy & Type & Observation & Dimension \\
        \specialrule{.15em}{.05em}{.05em} \\

        Locomotion Low-Level
        & Proprioceptive & Velocity command & 3 \\
        & & Projected gravity & 3 \\
        & & Body linear velocity & 3 \\
        & & Body angular velocity & 3 \\
        & & Joint position (excluding 4 wheel joints) & 12 \\
        & & Joint position error (excluding 4 wheel joints) for 3 half-LL time steps & 3 * 12 \\
        & & Joint velocity for 3 half-LL time steps & 3 * 16 \\
        & & Joint position target for 2 LL time steps & 2 * 16 \\ [2mm]
        & Exteroceptive & foot height scan around 4 feet & 4 * 52 \\

        \specialrule{.15em}{.05em}{.05em} \\

        \ac{MRMG} Navigation Mid-Level
        & Proprioceptive & Same as Low-Level & 140 \\ [2mm]
        & Exteroceptive & Body height scan & 600\\ [2mm]
        & Short Navigation State & Short goal position for 10 ML time steps & 10 * 6 \\

        \specialrule{.15em}{.05em}{.05em} \\

        \ac{MRMG} Navigation High-Level
        & Proprioceptive & Same as Low-Level & 140 \\ [2mm]
        & Exteroceptive & Same as Mid-Level & 600 \\ [2mm]
        & Short Navigation State & Same as Mid-Level & 60 \\ [2mm]
        & Common Neighbor States (GEE) & Neighbor velocity command for 20 HL time steps & 20 * 3 * \#N \\
        & & Neighbor position for 20 HL time steps & 20 * 3 * \#N \\
        & & Neighbor orientation for 20 HL time steps & 20 * 4 * \#N \\
        & & Neighbor linear velocity for 20 HL time steps & 20 * 3 * \#N \\
        & & Neighbor angular velocity for 20 HL time steps & 20 * 3 * \#N \\
        & & Neighbor projected gravity for 20 HL time steps & 20 * 3 * \#N \\ [2mm]
        & Task Neighbor State (GEE) & Neighbor short goal for 5 HL time steps & 5 * 6 * \#N \\ [2mm]
        & Long Navigation State (GEE)  & Long Goal position for 20 HL time steps & 20 * 3 * \#G \\
        & & Long Goal reach mask (0: not reached; 1: reached) & 1 * \#G \\
        
        \specialrule{.15em}{.05em}{.05em} \\
        
        Box Packing High-Level
        & Proprioceptive & Same as Low-Level & 140 \\ [2mm]
        & Exteroceptive & Same as Mid-Level & 430 \\ [2mm]
        & Common Neighbor States (GEE) & Same as above & 20 * 19 * \#N \\
        & Box State (GEE) & Box position for 20 HL time steps & 20 * 3 * \#B \\
        & & Box in-range mask (0: outside packing range; 1: inside packing range) & 1 * \#B \\
        
        \specialrule{.15em}{.05em}{.05em} \\
        
        Soccer High-Level
        & Proprioceptive & Same as Low-Level & 140 \\ [2mm]
        & Exteroceptive & Same as Mid-Level & 430 \\ [2mm]
        & Teammate State (GEE) & Common neighbor states & 20 * 19 * \#T \\
        & & team mask: 0: teammate; 1: opponent & 1 * \#N \\ [2mm]
        & Opponent State (GEE) & Common neighbor states & 20 * 19 * \#O \\
        & & team mask: 0: teammate; 1: opponent & 1 * \#N \\ [2mm]
        
        \specialrule{.15em}{.05em}{.05em} \\
        \multicolumn{4}{l}{\small \#N: number of neighbors. \#G: number of goals. \#B: number of boxes. \#T: number of teammates. \#O: number of opponents.} \\
        \multicolumn{4}{l}{\small GEE: observation goes through the \ac{GEE}} \\
    \end{tabular}
\end{table*}

}
\newpage
\vfill

\end{document}